\documentclass[10pt,twocolumn,letterpaper]{article}

\usepackage[pagenumbers]{cvpr} 

\usepackage{multirow}
\usepackage{enumitem}
\usepackage{fontawesome}
\usepackage{algorithm}
\usepackage{algorithmic}
\usepackage{graphicx}
\usepackage{caption}
\usepackage{subcaption}
\usepackage{makecell} 
\usepackage{placeins}
\usepackage{stfloats}
\usepackage{booktabs}
\usepackage{xcolor}
\usepackage{url}

\definecolor{cvprblue}{rgb}{0.21,0.49,0.74}
\usepackage[pagebackref,breaklinks,colorlinks,allcolors=cvprblue]{hyperref}

\def\paperID{11317} 
\def\confName{CVPR}
\def\confYear{2026}

\title{HarmoCLIP: Harmonizing Global and Regional Representations in Contrastive Vision-Language Models }

\author{
    Haoxi Zeng\textsuperscript{1} \quad 
    Haoxuan Li\textsuperscript{2} \quad 
    Yi Bin\textsuperscript{1}\thanks{Corresponding author.} \quad 
    Pengpeng Zeng\textsuperscript{1} \quad 
    Xing Xu\textsuperscript{1} \quad 
    Yang Yang\textsuperscript{2} \quad 
    Heng Tao Shen\textsuperscript{1} \\[2mm] 
    \textsuperscript{1}Tongji University \qquad 
    \textsuperscript{2}University of Electronic Science and Technology of China \\[2mm]
}

\begin{document}
\maketitle

\begin{abstract}
Contrastive Language-Image Pre-training (CLIP) has demonstrated remarkable generalization ability and strong performance across a wide range of vision-language tasks. However, due to the lack of region-level supervision, CLIP exhibits limited fine-grained semantic understanding. Although several methods attempt to mitigate this issue, they unintentionally disrupt the global alignment, resulting in a persistent trade-off where improving local perception simultaneously degrades global coherence. In this paper, we propose \textbf{HarmoCLIP}, a novel framework designed to harmonize global and region representations within CLIP. We first identify that the absence of direct alignment between local textual and visual semantics is the fundamental cause of the trade-off. To address this, HarmoCLIP introduces an explicit fine-grained semantic supervision term that directly aligns textual segments with their corresponding visual regions, effectively bridging the image region space and the textual space. To further strengthen the representation capability at the local level, our method introduces a novel Region–Language Alignment supervision strategy that promotes fine-grained semantic learning without compromising global semantic consistency. Extensive experiments demonstrate that HarmoCLIP achieves \textbf{state-of-the-art} (improvement up to \textbf{69.78\%}) performance on the global task of retrieval and yields a substantial \textbf{3.2\%} improvement in Top-1 accuracy on the region task of bounding-box classification, consistently outperforming prior approaches while providing a balanced, efficient, and plug-and-play solution to the global–local trade-off in CLIP. Code is available at \url{https://github.com/Erosist/HarmoCLIP}.
\end{abstract}
\section{Introduction}
\label{sec:intro}

\begin{figure}[t]
    \centering
    \includegraphics[width=\linewidth]{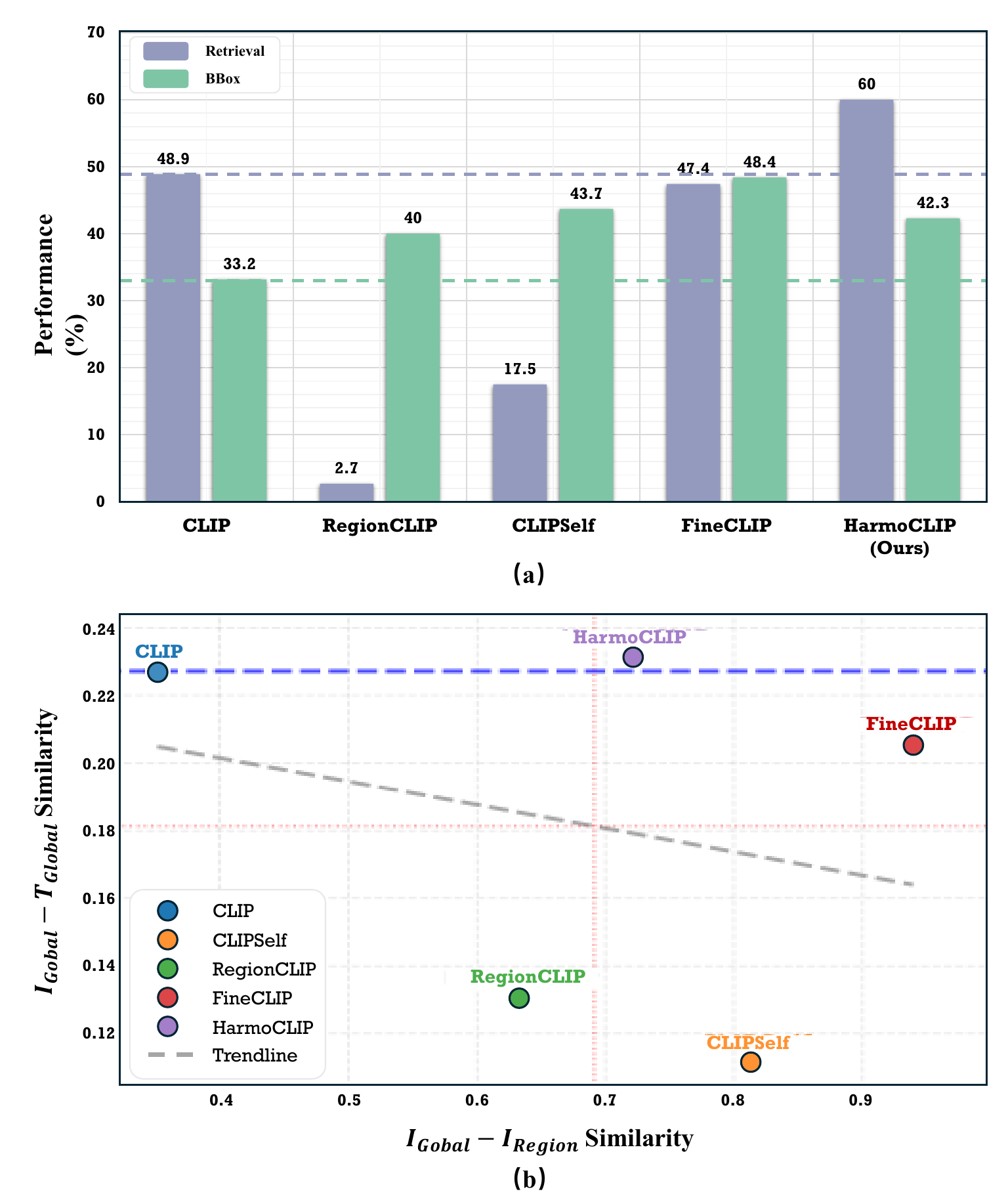}
    \caption{\textbf{Comparison of existing methods on global-awareness and region-level tasks.}
    Figure (a) shows the trade-off between global and region-level understanding across models. Figure (b) presents the relationship between the similarities of $I_{\text{R}}$-$I_{\text{G}}$ and $I_{\text{R}}$-$T_{\text{G}}$. Retrieval performance is reported as the mean of I$\rightarrow$T@1 and T$\rightarrow$I@1 on \emph{MSCOCO}, and BBox performance is measured by the Top-1 accuracy on \emph{OVCOCO}.}
    \label{fig:model_comparison}
\end{figure}

Vision-language representation learning~\cite{chen2020uniter,radford2021learning,li2020unicoder} emerges as a pivotal area within multimodal learning, underpinning a wide array of applications such as image classification~\cite{wu2023cora,recht2019imagenet,jia2021scaling}, cross-modal retrieval~\cite{bin2023unifying,pan2023prior,li2023your,zhang2024long,sun2024alpha}, and open-vocabulary detection~\cite{li2022blip,zareian2021open,zhou2022detecting,kim2023region}.
Contrastive Language-Image Pre-training (CLIP)~\cite{radford2021learning} stands as a foundational milestone in this domain, achieving remarkable success by learning a shared semantic space on large-scale image-text pairs through contrastive learning.
Due to its robust performance and generalization capabilities, CLIP has become a widely adopted backbone model in both academic research and industry applications~\cite{li2021supervision,gao2022pyramidclip}.

However, despite its widespread adoption, the standard CLIP exhibits two key limitations: \textbf{(1) Task–domain shift}~\cite{yao2023detclipv2}, where aligning entire images with full sentences hampers the ability of the model to focus on local regions or sub-image semantics~\cite{paiss2023teaching,ranasinghe2023perceptual,wu2024lotlip,zhong2022regionclip}; and \textbf{(2) Semantic asymmetry}, where the textual descriptions in the training data are often too brief to fully capture or match the detailed semantics contained in complete images~\cite{jing2024fineclip,zhou2022extract,fan2023improving}.
Consequently, the model demonstrates a lack of fine-grained semantic understanding.
This limitation becomes particularly evident when CLIP is deployed in region-level downstream tasks, such as bounding-box classification~\cite{wu2023cora} and open-vocabulary detection~\cite{li2022blip}.

To overcome these issues, recent works~\cite{zhong2022regionclip,wu2024clim,wu2023aligning,jing2024fineclip} have focused on enhancing the fine-grained understanding capabilities of CLIP, as illustrated in Figure~\ref{fig:model_comparison} (a). 
These studies either use extra data for region-level supervision~\cite{zhong2022regionclip} or enhance region-aware learning within the global semantic space~\cite{wu2023clipself}.
While these approaches do improve local perception, they often distort or even compromise the globally aligned text–image semantic space.
This results in a persistent trade-off between global awareness and fine-grained understanding.

In this work, we argue that existing methods, while striving to enhance fine-grained understanding, unintentionally introduce excessive alignment between global and region visual semantics, as illustrated in Figure~\ref{fig:model_comparison} (b), which leads to a degradation of global-level semantic consistency. To establish a more robust and interpretable connection between global–region and global–text semantic spaces, we first analyze the intrinsic causes of this trade-off, identifying that the instability of existing region–language alignment methods originates from their indirect alignment between local visual and textual semantic spaces, which ultimately disrupts the contrastive learning process and leads to the trade-off. Building upon this analysis, we first propose a novel region-language alignment strategy to enhance the representational capacity of the image region space.
Furthermore, to achieve direct and semantically coherent alignment between visual and textual representations, HarmoCLIP introduces a local-to-local supervision mechanism, aligning textual segments directly with corresponding visual regions. As a result, HarmoCLIP not only preserves the original global contrastive effectiveness of CLIP but also substantially strengthens fine-grained perception. Extensive experiments demonstrate that HarmoCLIP achieves balanced and consistent improvements across both image-level and region-level benchmarks, attaining state-of-the-art performance without relying on additional annotations, synthetic captions, or architectural modifications on overall performance, especially in cross-modal retrieval. HarmoCLIP is data-efficient and computationally efficient, providing general improvements for RLA-based methods and demonstrating strong potential for future multimodal learning frameworks. Our key contributions are as follows:
\begin{itemize}
\item We identify and empirically demonstrate the fundamental cause of the trade-off between local and global semantic alignment in CLIP, revealing that existing approaches rely on indirect alignment through intermediate semantic spaces built by global contrastive learning rather than direct region-to-text correspondence.
\item We introduce HarmoCLIP, a data-efficient training method that incorporates a novel RLA mechanism and directly aligns textual region-level inputs with image region-level features, significantly mitigating the trade-off between global and local semantics faced by existing CLIP fine-tuning strategies.
\item We conduct extensive evaluations on both global and region downstream benchmarks, and demonstrate that HarmoCLIP not only balances performance across global and region levels, showing notable improvement, but also reaches \textit{SOTA} in overall performance and retrieval.
\end{itemize}

\section{Related Works}
\label{sec:formatting}

\subsection{Contrastive Vision-Language Models}

The pursuit of vision–language alignment in vision–language models (VLMs)~\cite{wang2022multi,yao2023detclipv2} has been primarily driven by contrastive learning, which enables the acquisition of shared semantic representations across modalities. Early works such as ALIGN~\cite{jia2021scaling} and LiT~\cite{zhai2022lit} demonstrated the scalability and effectiveness of contrastive objectives for aligning large-scale visual and textual features under weak or noisy supervision. Building on this foundation, CLIP~\cite{radford2021learning} applied contrastive learning to billion-scale image–text pairs, achieving exceptional generalization in downstream tasks like retrieval~\cite{urbanek2024picture,chen2024sharegpt4v} and classification~\cite{deng2009imagenet,recht2019imagenet} through dual encoders that project visual and textual inputs into a unified embedding space. Following CLIP, methods including BLIP~\cite{li2022blip} and FLORENCE~\cite{yuan2021florence} further advanced this paradigm by incorporating multimodal pre-training and grounding mechanisms, collectively underscoring the versatility of contrastive learning for robust representation.

\subsection{Fine-Grained Understanding}

Despite the remarkable success of CLIP and related contrastive vision–language models, recent studies~\cite{yao2023detclipv2,jing2024fineclip,dong2023maskclip} have highlighted their limitations in fine-grained visual–textual alignment. CLIP often fails to establish strong correspondence between visual regions and textual descriptions~\cite{paiss2023teaching,ranasinghe2023perceptual}, mainly because its contrastive objective emphasizes global image–text alignment and the training captions are typically short and coarse~\cite{fan2023improving}. To overcome this issue, methods such as RegionCLIP~\cite{zhong2022regionclip} and CLIPSelf~\cite{wu2023clipself} introduce region-level alignment (RLA) mechanisms through region-specific text generation or distillation to encourage local feature learning, though these often compromise global semantic consistency. More recent efforts, including FineCLIP~\cite{jing2024fineclip} and FG-CLIP~\cite{xie2025fg}, leverage large language or vision–language models~\cite{li2023blip,alayrac2022flamingo} to generate fine-grained region-aware annotations, achieving improved local perception while enriching training semantics.

\begin{figure}[t]
    \centering
    \includegraphics[width=\linewidth]{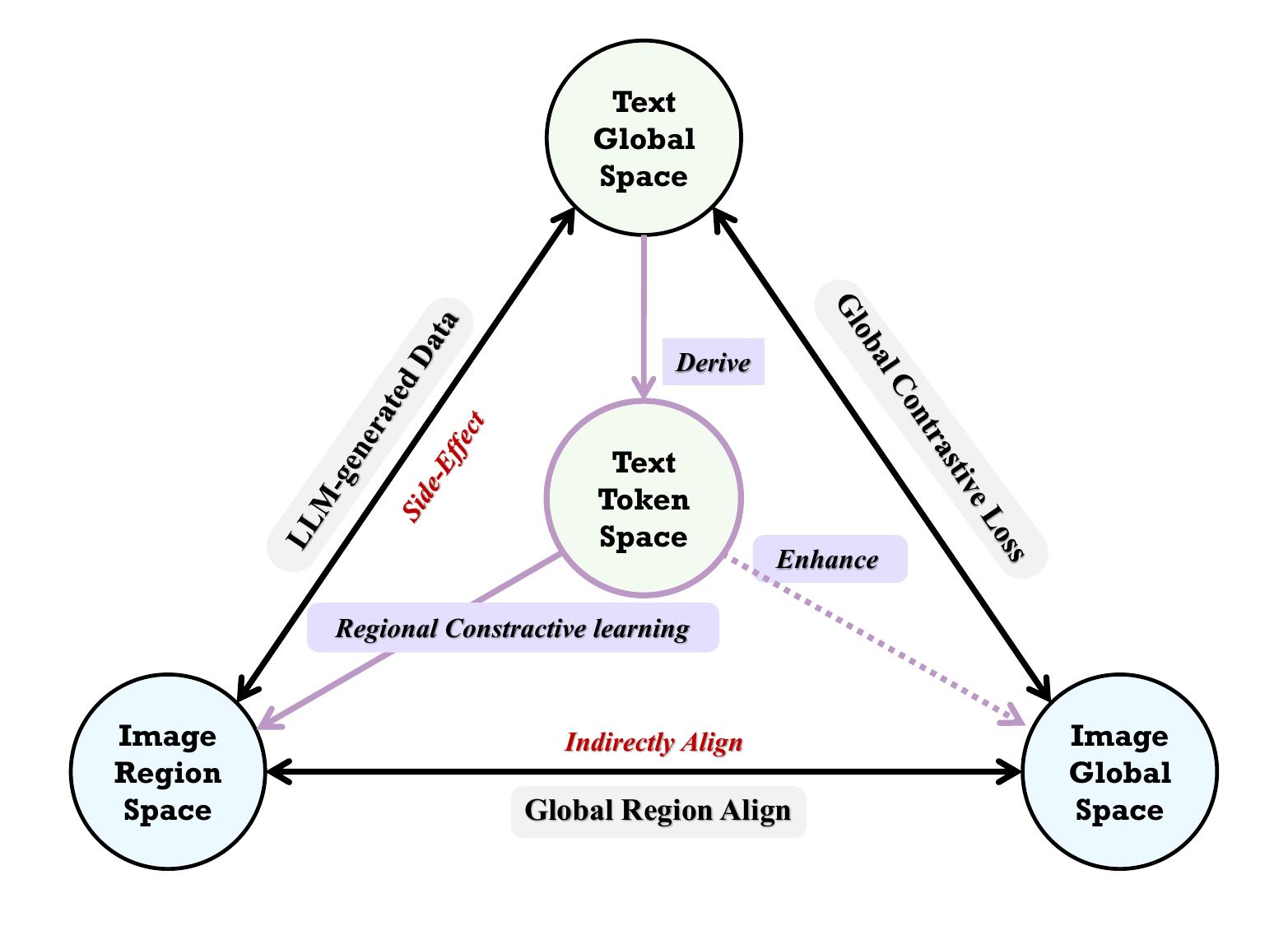}
    \caption{\textbf{An overview of the crucial semantic space in CLIP. }
\emph{Blue} arrows and annotations indicate the path of HarmoCLIP while \emph{Red} annotations show the limits of current methods.}
    \label{fig:semantic space}
\end{figure}

\section{Rethinking the Interplay Between Global and Region Semantics}
\label{relationship}

\subsection{CLIP Structure} 
CLIP employs a dual-encoder architecture that independently encodes images and text into a shared semantic space through contrastive learning~\cite{radford2021learning}.
On the visual side, the encoder divides each image into fixed-size patches, appends a [CLS] token and positional embeddings, and processes them through a transformer to obtain a global embedding that captures comprehensive semantic information.
By omitting the final transformer block, a dense feature map can be extracted~\cite{he2017mask,zhong2022regionclip}, which preserves abundant spatial and structural details within the image.
Region-level features are subsequently derived from these dense maps using RoI aggregation~\cite{he2017mask}, providing localized representations that complement the global embedding.
However, due to the hierarchical abstraction gap inherent in representation space~\cite{jing2024fineclip} of CLIP, these region features remain partially inconsistent with the global embedding in terms of semantic granularity.
On the text side, the encoder produces the text embedding from the projected representation of the end-of-text (EOT) token~\cite{devlin2019bert}, while intermediate token hidden states preserve contextual semantics corresponding to specific words.
This dual-encoder architecture naturally provides a foundation for learning region–text correspondences within our framework, enabling fine-grained semantic alignment between local visual regions and lexical units.

\begin{figure}[t]
    \centering
    \includegraphics[width=0.85\linewidth]{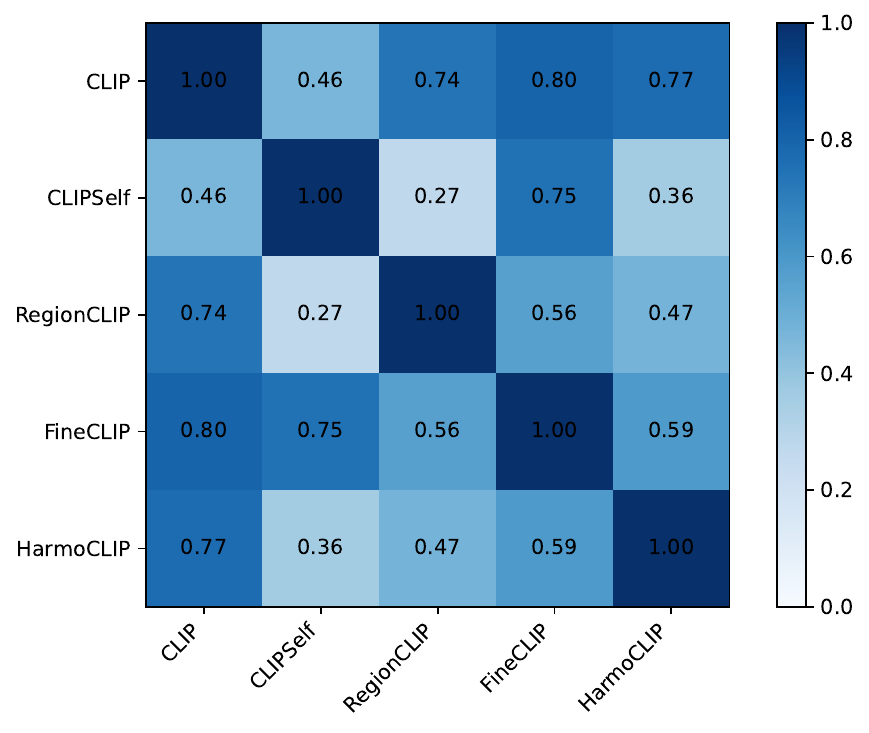}
    \caption{\textbf{\( I_{\text{G}} \)–\( I_{\text{R}} \) vs. \( I_{\text{R}} \)–\( T_{\text{G}} \) Concordance Matrix across models,} which reflects the strong correlation of two alignment processes.}
    \label{fig:concordance_matrix}

\vspace{0.1in}
\end{figure}

\subsection{Semantic Space Alignment}
The semantic space alignment in vision–language models~\cite{zhong2022regionclip,ranasinghe2023perceptual,radford2021learning} involves two primary directions: the alignment between the Image Global Space (\( I_{\text{G}} \)) and the Text Global Space (\( T_{\text{G}} \)), which corresponds to global cross-modal understanding such as retrieval tasks~\cite{young2014image,lin2014microsoft,liu2024mmbench}, and the alignment between the Image Region Space (\( I_{\text{R}} \)) and the Text Global Space (\( T_{\text{G}} \)), which reflects fine-grained perception abilities, such as bounding-box classification~\cite{zhong2022regionclip,wu2023clipself,jing2024fineclip}. From this perspective, CLIP employs Global Contrastive Learning to establish a strong alignment between \( I_{\text{G}} \) and \( T_{\text{G}} \). To achieve finer semantic correspondence between \( I_{\text{R}} \) and \( T_{\text{G}} \), methods such as RegionCLIP and FG-CLIP enhance the training data through prompt construction or VLM-generated captions. Meanwhile, CLIPSelf and FineCLIP rely on the pre-existing \( I_{\text{G}} \)–\( I_{\text{R}} \) alignment formed within CLIP, aligning \( I_{\text{R}} \) with \( I_{\text{G}} \) to achieve an indirect correspondence between \( I_{\text{R}} \) and \( T_{\text{G}} \). To provide a clearer illustration of this process, we visualize the relationships among these semantic spaces in Figure~\ref{fig:semantic space}.

However, we find that existing methods unintentionally reduce the semantic distance between global \( I_{\text{G}} \) and region \( I_{\text{R}} \) image representations. To further investigate the effect, we conduct a bounding-box classification task on the same batch of test data, comparing against the baseline CLIP. The results in Figure~\ref{fig:concordance_matrix} indicate that as models achieve a significant improvement in $I_{\text{R}}$-$T_{\text{G}}$ similarity, the semantic similarity of $I_{\text{R}}$-$I_{\text{G}}$ is highly improved as well, revealing a strong positive correlation. This phenomenon suggests that the $I_{\text{R}}$-$I_{\text{G}}$ alignment acts as an intermediary bridge in the process of enhancing fine-grained awareness. 

Given this correlation, we further analyze the interplay between the enhancement of the $I_{\text{R}}$-$I_{\text{G}}$ alignment and the $I_{\text{G}}$-$T_{\text{G}}$ learning process. Our results indicate that it is precisely this strengthening of the $I_{\text{R}}$-$I_{\text{G}}$ coupling that disrupts and degrades perception at the global level. In other words, the bridge formed by the $I_{\text{R}}$-$I_{\text{G}}$ alignment sacrifices a part of the essential $I_{\text{G}}$-$T_{\text{G}}$ alignment in exchange for fine-grained ability, as further evidenced in the supplementary material.

To summarize, we argue that the trade-off between global-level and region-level capabilities in CLIP primarily arises from the detrimental side-effect of this bridge. It induces an indirect alignment between \( I_{\text{R}} \) and \( T_{\text{G}} \), which in turn distorts the original semantic space of CLIP. Existing methods have largely overlooked this interplay, focusing solely on minimizing the \( I_{\text{R}} \)–\( T_{\text{G}} \) gap, which inadvertently results in performance degradation on global-level tasks, often even falling below the baseline. Therefore, the key challenge lies in developing a learning strategy that can construct a stronger and novel relationship between \( I_{\text{R}} \) and \( I_{\text{G}} \), while simultaneously enhancing the representational capacity of both the region-level and global-level semantic spaces. Therefore, based on the above analysis, we introduce a new semantic space, the \textit{Text Token Space}, and a novel supervision, the \textit{Lexeme-Region Contrastive Learning} to simultaneously reinforce the \( I_{\text{G}} \)–\( T_{\text{G}} \) alignment and establish a direct correspondence between \( I_{\text{R}} \) and \( T_{\text{G}} \).
\begin{figure*}[t]
    \centering
    \includegraphics[width=\linewidth]{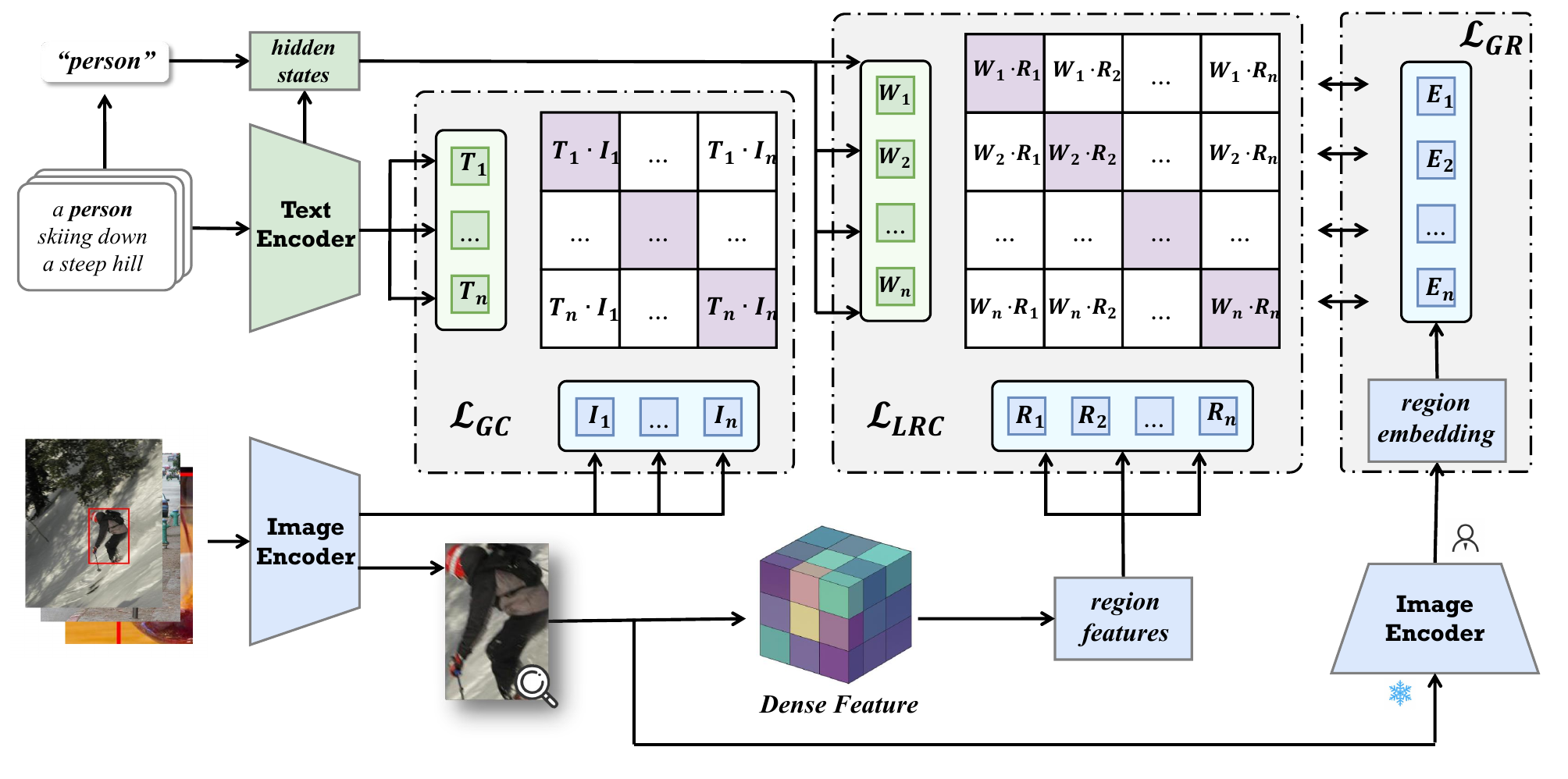}
    \caption{\textbf{Overall architecture of HarmoCLIP.} 
    It consists of three loss functions: 
    \( \mathcal{L}_{\mathrm{GC}} \) (Global Contrastive Learning), 
    \( \mathcal{L}_{\mathrm{LRC}} \) (Lexeme–Region Contrastive Learning), and 
    \( \mathcal{L}_{\mathrm{GR}} \) (Global-Region Alignment).}
    \label{fig:placeholder}
\end{figure*}

\section{Methodology}

\textbf{Overview.} Figure~\ref{fig:placeholder} presents the overall architecture of the proposed \textbf{HarmoCLIP}.
Built upon the original CLIP framework, HarmoCLIP further extends the standard Contrastive Learning paradigm by introducing the \textit{Global-Region Alignment} strategy and \textit{Lexeme-Region Contrastive Learning}.
For the inputs, beyond the conventional image--text pairs $\{I_i, T_i\}$, we incorporate grounding annotations to enrich them into region--word pairs $\{R_i, W_i\}$, where each word token $W_i$ semantically corresponds to a specific region crop $R_i$.
The pairs $\{I_i, T_i\}$ are fed through the encoders to construct a global contrastive alignment between the image and text embeddings. Meanwhile the pairs $\{R_i, W_i\}$ are processed to obtain region features from the image encoder and word hidden states from the text encoder for \textit{Lexeme-Region Contrastive Learning}.
On the visual side, we introduce a novel region-language alignment mechanism by leveraging a frozen CLIP model to extract the embeddings of cropped image regions. The region features from the trainable encoder are then aligned with frozen region embeddings, which reinforces the region-level semantic representations and enhances local perception without disrupting the original global semantic space.

\subsection{Global Contrastive Learning}
To preserve the original \( I_{\text{G}} \)–\( T_{\text{G}} \) alignment of CLIP and prevent semantic space drift during fine-tuning, we follow CLIP to perform Global Contrastive Learning. For each image–text pair $\{I_i, T_i\}$, we assume a one-to-one correspondence based on the index $i$ and obtain the text and image embeddings \( v_i, t_i \in \mathbb{R}^d \) through their respective encoders.

Within a batch of $N$ pairs, we measure the similarity between embeddings using the cosine similarity metric:
{\abovedisplayskip=5pt plus 2pt minus 2pt
 \belowdisplayskip=5pt plus 2pt minus 2pt
 \begin{equation}
 s(v_i, t_j) = \frac{v_i^\top t_j}{\lVert v_i\rVert \,\lVert t_j\rVert},
 \label{eq:cosine_similarity}
 \end{equation}}
where $s(v_i, t_j)$ denotes the similarity score between the $i$-th text and $j$-th image embedding.
Following the standard InfoNCE objective~\cite{he2020momentum}, we maximize the similarity of matched pairs while minimizing that of mismatched ones.
The loss function for a batch of $N$ image–text pairs is defined as:

\begin{equation}
\label{eq:global_contrastive_loss}
\begin{aligned}
\mathcal{L}_{\mathrm{GC}}
&= -\frac{1}{2N} \sum_{i=1}^{N}
\Bigg[
\log
\frac{
    \exp\!\big(s(v_i, t_i)/\tau\big)
}{
    \sum\limits_{j=1}^{N} \exp\!\big(s(v_i, t_j)/\tau\big)
}
\\[-3pt]
&\qquad\quad
+
\log
\frac{
    \exp\!\big(s(t_i, v_i)/\tau\big)
}{
    \sum\limits_{j=1}^{N} \exp\!\big(s(t_i, v_j)/\tau\big)
}
\Bigg],
\end{aligned}
\end{equation}
where $\tau$ is a learnable temperature parameter.

\subsection{Bridging Visual Regions and Textual Lexemes}
The Global Contrastive Learning framework of CLIP not only equips the model with cross-modal global representation capabilities but also establishes a bridge that enables alignment between the image region space and the text global space. However, as discussed in \textit{Sections~3.2}, methods without strong data augmentation often rely on this Global Contrastive Learning process to achieve indirect alignment between image regions and textual semantics by aligning \( I_{\text{R}} \) with \( I_{\text{G}} \). To address this, approaches such as FG-CLIP and FineCLIP leverage vision–language models (VLMs) to generate textual descriptions for image regions, attempting to solve the problem from a data perspective. Nevertheless, these approaches are computationally expensive and fail to address, at the technical level, the fundamental issue of balancing fine-grained perception and global semantic awareness.
Consequently, they still depend on the existing alignment space \( I_{\text{G}} \)–\( T_{\text{G}} \) and rely on the bridge formed by \( I_{\text{G}} \)–\( I_{\text{R}} \) to achieve indirect alignment, further emphasizing the dependence on this alignment path.

Moreover, as illustrated in Figure~\ref{fig:semantic space}, this bridge, while enhancing fine-grained perception, inadvertently disrupts the Global Contrastive Learning process of \( I_{\text{G}} \)–\( T_{\text{G}} \). In other words, during the indirect alignment process, relying on the Global Contrastive Learning bridge sacrifices global semantic consistency in exchange for localized detail. This imbalance ultimately leads to a loss of global awareness. 

To overcome this limitation, a new bridge is urgently needed that can facilitate alignment between the image region space and the text global space while preserving the integrity of the Global Contrastive Learning process. Motivated by this, we extend the original text global space into a finer-grained \textit{Text Token Space}, enabling direct local-to-local alignment between words in captions and image regions. This new alignment strategy, termed \textbf{Lexeme–Region Contrastive Learning $\mathcal{L}_{\mathrm{LRC}}$}, effectively replaces the original dependence on \( I_{\text{G}} \)–\( I_{\text{R}} \) coupling with a more stable and semantically consistent pathway.

To perform Lexeme–Region Contrastive Learning, we first obtain the hidden states of the word pieces and the corresponding region features from each paired sample $\{W_i, R_i\}$. For extracting word-piece hidden states, we remove the final projection layer of the text encoder~\cite{devlin2019bert}. From the transformer output sequence of hidden states $(h_1, h_2, \dots, h_L)$, where $h_i \in \mathbb{R}^{B \times d_{\mathrm{model}}}$, we locate the target word corresponding to a given region $R_i$ as follows:
{\abovedisplayskip=5pt plus 2pt minus 2pt
 \belowdisplayskip=5pt plus 2pt minus 2pt
 \begin{equation}
 W_{\mathrm{emb}}
 = h^{T} = (h_1,\dots,h_L).
 \label{eq:word_embedding_1}
 \end{equation}}
{\abovedisplayskip=5pt plus 2pt minus 2pt
 \belowdisplayskip=5pt plus 2pt minus 2pt
 \begin{equation}
 W_{\mathrm{target}}
 = W_{\mathrm{emb}}[:,\,\mathrm{token\_idx},:]
 = h_{\mathrm{token\_idx}}.
 \label{eq:word_embedding_4}
 \end{equation}}

To obtain fine-grained region features, it is essential to bypass the global aggregation mechanism of the Vision Transformer, where the final self-attention operation fuses patch-level information and thus weakens local details. We leverage the rich localized semantics contained in the Value features ($V$) before this aggregation. Following MaskCLIP~\cite{zhou2022extract}, we remove the self-attention operation in the final Transformer block while retaining the projection layers, layer normalizations, and FFNs. 
In the final residual block of the image encoder, the modified residual attention operation can be expressed as:
{\abovedisplayskip=5pt plus 2pt minus 2pt
 \belowdisplayskip=5pt plus 2pt minus 2pt
 \begin{equation}
 \label{eq:modified_resattn}
 \begin{aligned}
 z' &= \operatorname{ModifiedResAttn}(x), \\ 
 y' &= x + \operatorname{Proj}(v), \\
 z' &= y' + \operatorname{FFN}(y').
 \end{aligned}
 \end{equation}}
This modification enables the model to reshape the output image embeddings $z'[1:h\times w]$ into a spatially structured feature map of size $h\times w$, representing dense region visual information.
On top of this dense feature map, we apply RoIAlign~\cite{he2017others} to extract region-specific representations for each bounding box, obtaining the region feature $R_{\text{target}}$ as:
{
 \begin{equation}
 \label{eq:roi_align}
 R_{\text{target}} = \operatorname{RoIAlign}(\mathrm{Image}_{\text{dense}}).
 \end{equation}}
 
 Similar to Global Contrastive Learning, for a batch size of N, after obtaining the region–text pairs $\{R_i, W_i\}$ and their corresponding feature representations $\{r_i, l_i\}$, we compute the region contrastive loss as follows:

\begin{equation}
\label{eq:region_contrastive_loss}
\begin{aligned}
\mathcal{L}_{\mathrm{LRC}}
&= -\frac{1}{2N} \sum_{i=1}^{N}
\Bigg[
\log
\frac{
    \exp\!\big(s(r_i, l_i)/\tau\big)
}{
    \sum\limits_{j=1}^{N} \exp\!\big(s(r_i, l_j)/\tau\big)
}
\\[-3pt]
&\qquad\quad
+
\log
\frac{
    \exp\!\big(s(l_i, r_i)/\tau\big)
}{
    \sum\limits_{j=1}^{N} \exp\!\big(s(l_i, r_j)/\tau\big)
}
\Bigg].
\end{aligned}
\end{equation}

\begin{table*}[t]
  \caption{\textbf{Overall comparison across different CLIP-based models.} 
  RegionCLIP~\cite{zhong2022regionclip} uses RN50 as backbone while all others use ViT-B/16.
  $^{\dagger}$ indicates methods trained with VLM-generated data or larger-scale datasets.
  Bold numbers denote the best performance among methods without data enhancement.}
  \label{tab:overall_comparison}
  \centering
  \setlength{\tabcolsep}{6pt}
  \begin{tabular}{@{}l l ccccc|cc@{}}
    \toprule
    \textbf{Task} &
    \textbf{Metric} 
    & \textbf{CLIP}
    & \textbf{EVA-CLIP}
    & \textbf{RegionCLIP}
    & \textbf{CLIPSelf}
    & \textbf{HarmoCLIP}
    & \textbf{FineCLIP$^{\dagger}$}
    & \textbf{FG-CLIP$^{\dagger}$} \\
    \midrule

    \multirow{2}{*}{\textbf{Retrieval}} 
    & \textbf{I$\rightarrow$T@1}
    & 51.80 & 57.32 & 3.40 & 18.80 & \textbf{69.78} & 54.50 & 64.10 \\
    & \textbf{T$\rightarrow$I@1}
    & 37.60 & 40.47 & 2.00 & 16.10 & \textbf{53.44} & 40.20 & 45.40 \\

    \midrule

    \multirow{2}{*}{\textbf{BBox}}
    & \textbf{OVCOCO}
    & 31.10 & 30.60 & 40.00 & \textbf{43.70} & 43.20 & 48.40 & 52.30 \\
    & \textbf{LVIS}
    & 20.90 & 14.40 & 22.20 & 7.80 & \textbf{22.20} & 23.30 & 28.60 \\

    \midrule

    \textbf{Sum} & -- 
    & 141.40 & 142.79 & 67.60 & 86.40 & \textbf{188.62} & 166.40 & 190.40 \\
    \bottomrule
  \end{tabular}
\end{table*}

\subsection{Global-Region Alignment}
After establishing an appropriate alignment path between the image region space and the text global space, it becomes crucial to enhance region-level awareness by strengthening the representation of the image region space while achieving effective alignment. In existing contrastive training, region features are only treated as implicit components of the global representation and thus receive no direct supervision, resulting in a lack of detailed and explicit representation within the image region space. Therefore, methods such as RegionCLIP~\cite{zhong2022regionclip} and FineCLIP~\cite{jing2024fineclip} introduce a region–language alignment strategy to enhance the image region space by reducing the distance between \( I_{\text{R}} \) and \( I_{\text{G}} \). However, these approaches overlook the issue of semantic asymmetry: since the textual semantic space is inherently discontinuous and carries less information than the visual space, directly aligning it with \( I_{\text{R}} \) tends to introduce noise and degrade the quality of region representations. To address this problem, we propose a Global–Region Alignment Loss that leverages the richer and more stable image global space to provide effective supervision, as described below.

Subsequently, we employ a frozen CLIP encoder to process the region crops, producing richer region embeddings denoted as $\{E_i\}$.
These embeddings serve as teacher representations, while the learned region features are aligned with them.
The alignment objective is defined as:
{\abovedisplayskip=5pt plus 2pt minus 2pt
 \belowdisplayskip=5pt plus 2pt minus 2pt
 \begin{equation}
 \label{eq:stage_loss}
 \mathcal{L}_{\text{GR}}
 = 1 - \frac{1}{B}
 \sum_{i=1}^{B}
 \frac{R_i}{\lVert R_i \rVert_2}
 \cdot
 \frac{E_i}{\lVert E_i \rVert_2},
 \end{equation}}
where $B$ denotes the batch size.

\subsection{Total Learning Objective}
Our proposed HarmoCLIP integrates the three key components described above, namely Global Contrastive Learning ($\mathcal{L}_{\mathrm{GC}}$), Global–Region Alignment ($\mathcal{L}_{\mathrm{GR}}$), and Lexeme–Region Contrastive Learning ($\mathcal{L}_{\mathrm{LRC}}$), into a unified optimization framework. The overall learning objective is formulated as:

{\abovedisplayskip=5pt plus 2pt minus 2pt
 \belowdisplayskip=5pt plus 2pt minus 2pt
 \begin{equation}
 \label{eq:total_loss}
 \mathcal{L}
 = \mathcal{L}_{\mathrm{GC}}
 +  \mathcal{L}_{\mathrm{LRC}}
 +  \mathcal{L}_{\mathrm{GR}}.
 \end{equation}}
Together, these components jointly optimize both global and local semantics in a coherent manner.

\section{Experiment}

\subsection{Experiment Details}
We train HarmoCLIP on a single L40 GPU with a batch size of 50 and 8 data-loading workers.
The model is optimized for 5 epochs using AdamW~\cite{loshchilov2017decoupled} with a learning rate of $1\times10^{-5}$ and a weight decay of $0.1$. A cosine annealing schedule with a 1000-step warm-up is applied.

For training data, we merge the COCO2017 Captions and COCO2017 Instances datasets~\cite{lin2014microsoft}.
For each image, its caption–image and image–annotation pairs are combined to form unified samples containing both caption–image and word–region correspondences, resulting in 599K annotated pairs across 95K images.

All experiments are conducted with ViT-B/16 and ViT-L/14 backbones from EVA-CLIP~\cite{sun2023eva}, selected for their balance between efficiency and representational strength.
Using the COCO \texttt{val2017} split, we evaluate HarmoCLIP on box classification and retrieval tasks based on pooled region features and global embeddings.
We report Top-1 and Top-5 mean accuracy for box classification, and Recall@1 (R@1) for both image-to-text and text-to-image retrieval.

\begin{table}[ht!]
  \caption{\textbf{Region-level evaluation results on OVCOCO and LVIS.} 
  We report Top-1 accuracy for zero-shot bounding-box classification. 
  $^{\dagger}$ denotes methods trained with additional data or VLM-based enhancement. 
  Bold indicates the best results without data enhancement. }
  \label{tab:region_eval}
  \centering
\setlength{\tabcolsep}{3pt}
\begin{tabular}{lcccc}
\toprule
\textbf{Method} & \multicolumn{2}{c}{\textbf{OVCOCO}} & \multicolumn{2}{c}{\textbf{LVIS}} \\
\cmidrule(lr){2-3} \cmidrule(lr){4-5}
\textbf{Backbone} & \textbf{ViT-B/16} & \textbf{ViT-L/14} & \textbf{ViT-B/16} & \textbf{ViT-L/14} \\
\midrule
CLIP        & 31.1  & 33.8  & 20.9  & 9.3   \\ 
EVA-CLIP    & 30.6  & 46.14 & 14.4  & 18.3  \\
RegionCLIP  & 40.0  & --    & \textbf{22.2} & --    \\
CLIPSelf    & \textbf{43.7} & --    & 7.8    & --    \\
\textbf{HarmoCLIP}   & 43.2  & \textbf{61.05} & \textbf{22.2} & \textbf{24.47} \\
\midrule
FineCLIP$^{\dagger}$    & 48.4 & 54.5 & 23.3 & 22.5 \\
FG-CLIP$^{\dagger}$     & 52.3 & 63.2 & 28.6 & 38.3 \\
\bottomrule
\end{tabular}
\end{table}

\subsection{Overall Performance Comparison}

As illustrated in Table~\ref{tab:overall_comparison}, we conducted a series of experiments designed to evaluate both the global and local representational capabilities of the model.
The evaluation covers two tasks: cross-modal retrieval on \textbf{MSCOCO 5K}~\cite{lin2014microsoft}, chosen for its larger scale, and bounding-box classification on \textbf{OVCOCO}~\cite{lin2014microsoft} and \textbf{LVIS}~\cite{gupta2019lvis}.
We directly report I$\rightarrow$T@1 and T$\rightarrow$I@1 for retrieval, and Top-1 Accuracy on both OVCOCO and LVIS for bounding-box classification. To more intuitively reflect the overall performance of each model, we additionally report a \textit{Sum} score that aggregates results across these evaluation tasks.
Since RegionCLIP and CLIPSelf do not provide ViT-L/16 variants, all models other than RegionCLIP use ViT-B/16 as the backbone.

As shown by the results, models such as CLIPSelf and RegionCLIP exhibit clear overfitting to fine-grained tasks, while FineCLIP~\cite{jing2024fineclip}, even with VLM-based data augmentation, still suffers from a decline in retrieval performance.
Under the same data scale, our proposed \textbf{HarmoCLIP} achieves the best overall performance and is the only method that surpasses the EVA-CLIP baseline across all tasks, demonstrating its ability to balance region understanding with global semantic awareness.

\subsection{Region-level Task}
\textbf{Bounding-Box Classification.}
To evaluate the capability of the model in recognizing fine-grained visual information, we perform \textbf{zero-shot bounding-box classification} on \textbf{OVCOCO}~\cite{lin2014microsoft} and \textbf{LVIS}~\cite{gupta2019lvis}, following the evaluation protocol described in FineCLIP~\cite{jing2024fineclip}. 
This downstream task is designed to assess the region-level awareness of the model in cross-modal scenarios and its capability for fine-grained object discrimination.

As shown in Table~\ref{tab:region_eval}, \textbf{HarmoCLIP} demonstrates enhanced fine-grained perception compared with models of the same category and achieves highly competitive overall performance across both metrics among methods without data enhancement, while still maintaining competitive accuracy relative to approaches such as FG-CLIP and FineCLIP that rely on larger scale of data from VLMs.

\begin{table}[ht]
  \caption{\textbf{Cross-modal retrieval performance on MSCOCO 5K and Flickr30K.} 
  $^{\dagger}$ marks models with extra data or VLM augmentation. 
  \textbf{Bold} denotes the best among standard methods.}
  \label{tab:retrieval}
  \centering
  \setlength{\tabcolsep}{6pt}
  \begin{tabular}{@{}lcccc@{}}
    \toprule
    \textbf{Method} &
    \multicolumn{2}{c}{\textbf{MSCOCO}} &
    \multicolumn{2}{c}{\textbf{Flickr30K}} \\
    \cmidrule(lr){2-3} \cmidrule(lr){4-5}
    & I$\rightarrow$T@1 & T$\rightarrow$I@1 & I$\rightarrow$T@1 & T$\rightarrow$I@1 \\
    \midrule
    CLIP & 51.80 & 37.60 & 82.20 & 62.10 \\
    EVA-CLIP & 57.32 & 40.47 & 85.70 & 71.20 \\
    RegionCLIP & 3.40 & 2.00 & 3.90 & 7.90 \\
    CLIPSelf & 18.80 & 16.10 & 33.80 & 35.00 \\
    \textbf{HarmoCLIP} & \textbf{69.78} & \textbf{53.44} & \textbf{90.80} & \textbf{76.46} \\
    \midrule
    FineCLIP$^{\dagger}$ & 54.50 & 40.20 & 82.50 & 67.90 \\
    FG-CLIP$^{\dagger}$ & 64.10 & 45.40 & 90.70 & 76.40 \\
    \bottomrule
  \end{tabular}
\end{table}

\subsection{Image-level Task}
\textbf{Cross-Modal Retrieval.}
To comprehensively evaluate the global semantic understanding of current methods, we conducted \textbf{Retrieval} experiments on two classic benchmarks: \textbf{MSCOCO 5K}~\cite{lin2014microsoft} and \textbf{Flickr30K}~\cite{young2014image}. 
As shown in Table~\ref{tab:retrieval}, \textbf{HarmoCLIP} achieves a substantial improvement over existing CLIP-based and variant models, attaining \textit{state-of-the-art} performance on both evaluation metrics. 

Notably, our model achieves these results without relying on VLM-generated data and with only limited training samples, highlighting its data-efficient and computationally efficient design.
These findings demonstrate that \textbf{HarmoCLIP} effectively aligns visual and textual representations within a unified semantic space, reflecting superior global-level understanding and fine-grained capacity.

\subsection{Ablation Study}
\label{ablation}

\textbf{Ablation of Objective Components.}
Our HarmoCLIP framework consists of three main objective components: 
the \textbf{Global Contrastive Loss} ($\mathcal{L}_{\mathrm{GC}}$), 
the \textbf{Lexeme-Region Contrastive Learning} ($\mathcal{L}_{\mathrm{LRC}}$), 
and the \textbf{Global-Region Alignment Loss} ($\mathcal{L}_{\mathrm{GR}}$). 
We begin by analyzing the baseline model, followed by the incremental effects of $\mathcal{L}_{\mathrm{GC}}$ and $\mathcal{L}_{\mathrm{LRC}}$.

As shown in \emph{Table~\ref{tab:ablation}} (row 2), using only $\mathcal{L}_{\mathrm{GC}}$, which follows the same optimization path as the base CLIP, ensures that the model learns strong global semantic understanding.
On top of this baseline, we sequentially introduce $\mathcal{L}_{\mathrm{LRC}}$ and $\mathcal{L}_{\mathrm{GR}}$ 
to examine their independent and combined effects.

In \emph{Table~\ref{tab:ablation}} (row 3), the inclusion of the Lexeme-Region Contrastive Loss 
($\mathcal{L}_{\mathrm{LRC}}$) significantly boosts retrieval performance—
from 57.32\% to 70.14\% on I$\rightarrow$T@1 
and from 40.47\% to 52.21\% on T$\rightarrow$I@1. 
This demonstrates that the region–text contrastive alignment strategy effectively enhances the global perception of the model. 

Further, we evaluate the impact of incorporating only the Global–Region Alignment Loss ($\mathcal{L}_{\mathrm{GR}}$). 
Notably, introducing $\mathcal{L}_{\mathrm{GR}}$ leads to a substantial improvement on the BBox classification task (from 33.21\% to 44.31\%), indicating that deliberate global–region alignment significantly enhances region-level semantic understanding. 
Meanwhile, the retrieval performance remains comparable to the baseline, demonstrating that our region–language alignment is both novel and efficient, which markedly distinguishes it from previous methods.

Finally, when all three objectives
$\mathcal{L}_{\mathrm{GC}}$ + $\mathcal{L}_{\mathrm{LRC}}$ + $\mathcal{L}_{\mathrm{GR}}$
are jointly optimized, the model achieves consistent improvements across both global and region-level tasks.
This demonstrates that incorporating $\mathcal{L}_{\mathrm{LRC}}$ and $\mathcal{L}_{\mathrm{GR}}$ effectively mitigates the trade-off and leads to well-balanced semantic spaces.
In contrast, models trained without either $\mathcal{L}_{\mathrm{LRC}}$ or $\mathcal{L}_{\mathrm{GR}}$ tend to show opposite trends between the two tasks, where the improvement of one usually results in a decline of the other.
In other words, these two objectives have a certain degree of trade-off when applied separately, but when optimized together, they complement each other and jointly enhance the overall performance.

\begin{table}[bt]
  \caption{\textbf{Ablation study on HarmoCLIP objective components.} 
  Each row reports performance when incrementally adding each objective term. 
  }
  \label{tab:ablation}
  \centering
  \setlength{\tabcolsep}{4.5pt} 
  \renewcommand{\arraystretch}{1.1} 
  \begin{tabular}{@{}lcc|cc@{}}
    \toprule
    \multirow{2}{*}{\textbf{Method}} & 
    \multicolumn{2}{c|}{\textbf{BBox}} & 
    \multicolumn{2}{c}{\textbf{Retrieval}} \\
    \cmidrule(lr){2-3} \cmidrule(lr){4-5}
    & \textbf{Top-1} & \textbf{Top-5} & \textbf{I$\rightarrow$T@1} & \textbf{T$\rightarrow$I@1} \\
    \midrule
    CLIP & 33.21 & 55.28 & 57.32 & 40.47 \\
    +$\mathcal{L}_{\mathrm{GC}}$ & 37.42 & 59.55 & 62.86 & 47.50 \\
    +$\mathcal{L}_{\mathrm{GC}}$+$\mathcal{L}_{\mathrm{LRC}}$ & 35.48 & 61.04 & 70.14 & 52.21 \\
    +$\mathcal{L}_{\mathrm{GC}}$+$\mathcal{L}_{\mathrm{GR}}$ & 44.31 & 70.36 & 57.74 & 44.90 \\
    +$\mathcal{L}_{\mathrm{GC}}$+$\mathcal{L}_{\mathrm{LRC}}$+$\mathcal{L}_{\mathrm{GR}}$ & 42.27 & 68.73 & 68.24 & 52.07 \\
    \bottomrule
  \end{tabular}
\vspace{-0.1in}
\end{table}

\textbf{Ablation of Region-Language Alignment Method.}
To further investigate the impact of the Lexeme-Region Contrastive Loss ($\mathcal{L}_{\mathrm{LRC}}$) on enabling the model to achieve a harmonious balance between global-level and region-level understanding, we conducted additional experiments by substituting the region–language alignment components in other representative methods. Specifically, we initialized models with CLIPSelf and RegionCLIP, then performed stage-2 fine-tuning using our proposed $\mathcal{L}_{\mathrm{LRC}}$.
As shown in Table~\ref{tab:llrc_results}, the models trained with $\mathcal{L}_{\mathrm{LRC}}$ for even a single epoch rapidly regained strong global semantic expressiveness while preserving the fine-grained perception previously achieved by the original methods. These results not only validate our theoretical analysis of the trade-off but also demonstrate that the proposed $\mathcal{L}_{\mathrm{LRC}}$ possesses plug-and-play flexibility, offering a general framework to mitigate the inherent tension between global awareness and local understanding in contrastive vision–language models.

\begin{table}[ht]
\centering
\caption{Performance comparison on image--text retrieval and BBOX classification tasks.}
\label{tab:llrc_results}
\begin{tabular}{@{}l c c@{}}
\toprule
\textbf{Method} & \textbf{Retrieval} & \textbf{BBOX Classification} \\
\midrule
CLIPSelf & 16.10 & 43.70 \\

CLIPSelf + $\mathcal{L}_{\mathrm{LRC}}$ &
46.11\rlap{\textsubscript{\textcolor{red}{+30.01}}} &
43.20\rlap{\textsubscript{\textcolor{red}{-0.50}}} \\

\midrule
RegionCLIP & 2.00 & 40.00 \\

RegionCLIP + $\mathcal{L}_{\mathrm{LRC}}$ &
47.51\rlap{\textsubscript{\textcolor{red}{+45.51}}} &
42.01\rlap{\textsubscript{\textcolor{red}{+2.01}}} \\
\bottomrule
\end{tabular}
\vspace{-0.1in}
\end{table}

\textbf{Ablation of Image Encoder Fine-Tuning Depth.}
In this section, we analyze how the fine-tuning depth of the image encoder affects the overall performance of HarmoCLIP, while the text encoder remains fully fine-tuned throughout all experiments. Following the design strategy used in CLIPSelf~\cite{wu2023clipself}, we vary the number of trainable transformer layers in the image encoder—specifically testing depths of 3, 6, 9, and 12 layers—based on the ViT-B/16 backbone, with the input image resolution fixed at $224 \times 224$.

As shown in Table~\ref{tab:layer_ablation}, the number of trainable layers in the image encoder significantly affects performance. Fine-tuning the last 12 layers yields the best balance between retrieval accuracy and region-level classification. This result is consistent with CLIPSelf, indicating that moderate tuning depth effectively adapts to region-level supervision while preserving pretrained global alignment.

\begin{table}[h]
  \caption{\textbf{Ablation on the number of unlocked layers in the image encoder.} 
  We compare different numbers of trainable layers while keeping other settings fixed.
  Metrics include Recall@1 for image-to-text (I$\rightarrow$T@1) and text-to-image (T$\rightarrow$I@1) retrieval,
  and Top-1/Top-5 accuracy for box classification.}
  \label{tab:layer_ablation}
  \centering
  \setlength{\tabcolsep}{5pt} 
  \begin{tabular}{@{}ccccc@{}}
    \toprule
    \textbf{Layers Unlocked} & \textbf{I$\rightarrow$T@1} & \textbf{T$\rightarrow$I@1} & \textbf{Top-1} & \textbf{Top-5} \\
    \midrule
    3  & 68.24 & 52.07 & 42.27 & 68.73 \\
    6  & 69.36 & 53.34 & 43.90 & 70.19 \\
    9  & 69.38 & 53.48 & 43.92 & 69.73 \\
    12 & 69.78 & 53.44 & 43.16 & 69.16 \\
    \bottomrule
  \end{tabular}
\vspace{-0.05in}
\end{table}

\section{Conclusion}
In this paper, we propose \textbf{HarmoCLIP}, a novel CLIP-based framework designed to balance local fine-grained perception and global semantic understanding within a unified model. Our work first identifies the underlying reasons why existing approaches fail to achieve this balance, and subsequently validate our findings through extensive experiments.
HarmoCLIP consistently achieves superior performance across multiple downstream tasks, notably setting a new \textit{state-of-the-art} in retrieval benchmarks while maintaining the best overall representational balance among competing methods. Our approach is efficient, and provides novel insights into resolving the long-standing trade-off between region sensitivity and global awareness in vision-language models.
{
    \small
    \bibliographystyle{ieeenat_fullname}
    \bibliography{main}
}

\clearpage
\maketitlesupplementary

\setcounter{table}{0}
\renewcommand{\thetable}{S\arabic{table}}
\setcounter{figure}{0}
\renewcommand{\thefigure}{S\arabic{figure}}

\section{Datasets details}
\begin{figure}[t]   
    \centering
    \includegraphics[width=\linewidth]{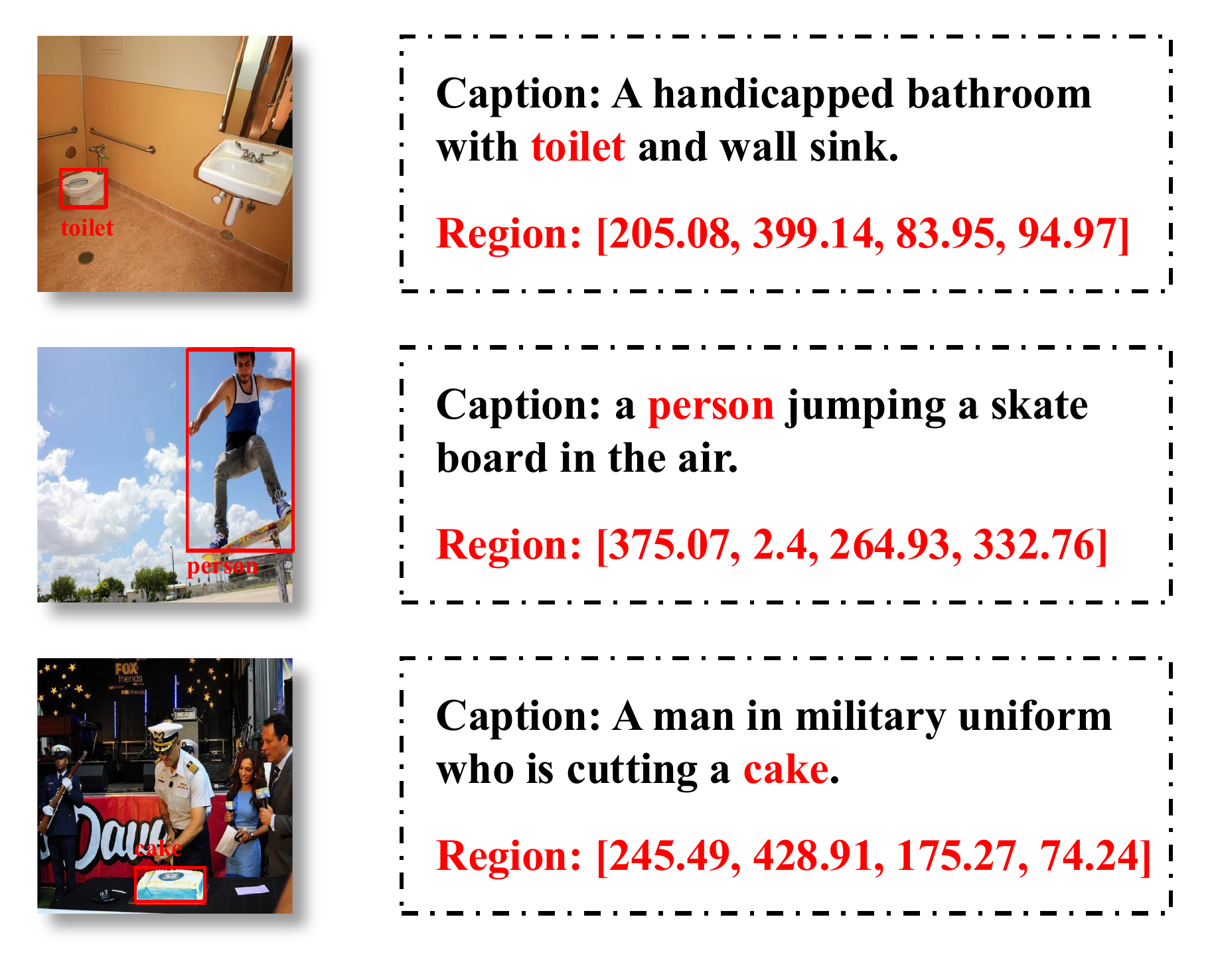}
    \caption{Data samples in details.}
    \label{fig:dataset}
\end{figure}

We merge the COCO2017 Captions and COCO2017 Instances datasets to construct a dataset that simultaneously provides cross-modal image–caption pairs and image–region annotations with corresponding words. In COCO2017, each image is associated with 4 to 5 captions, as well as 3 to 8 instance-level bbox annotations. However, many objects annotated by bounding boxes do not appear in the captions.

For each image, we therefore match the category label of each bbox annotation with the tokens in the caption. If no correspondence is found, that case is discarded. As a result, each caption in our dataset contains only a single word aligned with one region. When a caption contains multiple words that correspond to different instance annotations, we split it into multiple data samples.
Through this process, we construct 599K word-level annotations across 95K images.

To more clearly illustrate the composition and characteristics of our dataset, we provide several example instances in Figure~\ref{fig:dataset}. Each data sample consists of an image paired with its caption, which serves as the global text. Within each image, instance-level annotations provide bounding-box regions that are used as region crops. Each region crop is associated with a labeled word that semantically describes the region, and this word always appears within the corresponding caption.

\begin{figure}[t]
    \centering
    \includegraphics[width=\linewidth, clip]{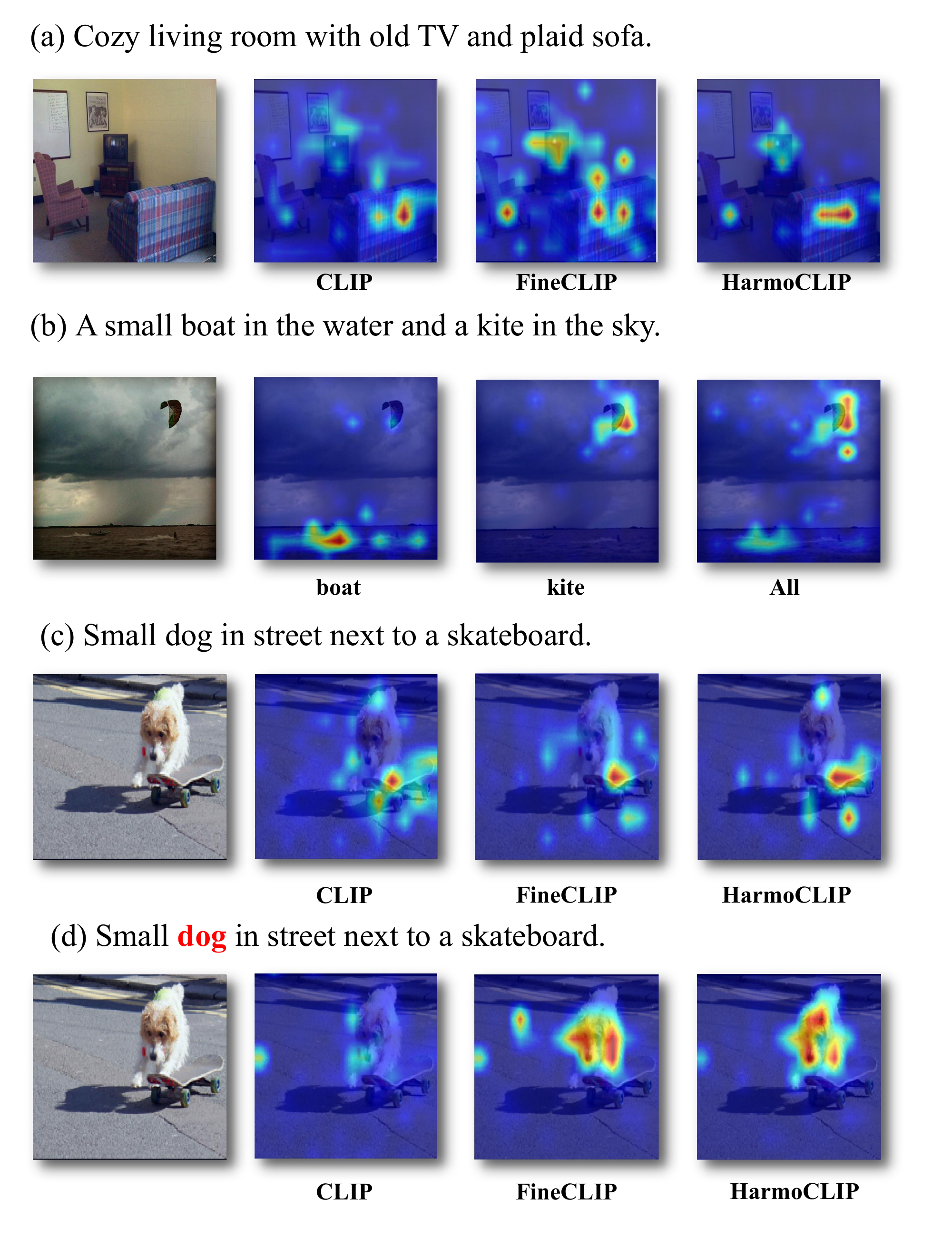}
    \caption{\textbf{Qualitative visualization of representations.}
    (a) shows global-level sentence visualizations across models, (b) display token-level attention maps of HarmoCLIP, and (c)–(d) illustrate global and region visualizations, demonstrating its capacity to capture both holistic and detailed semantics.}
    \label{fig:visualize_overall}
\end{figure}

\section{Visualization Results}
To comprehensively demonstrate the global and local semantic understanding across modalities of models, and to intuitively illustrate the improvements achieved by our method, we employ GAE visualization on sentence-level textual inputs. As shown in Figure~\ref{fig:visualize_overall}(a), the attention heatmaps of full sentences reveal the global-level attention distribution. Compared with other approaches, HarmoCLIP demonstrates finer global semantic capture and more accurate spatial localization, effectively addressing the limitations of previous models. Figure~\ref{fig:visualize_overall}(b) further highlights its strong fine-grained semantic awareness.
In Figures~\ref{fig:visualize_overall}(c)–(d), we visualize both sentence-level and token-level attentions to jointly assess global representational capacity and fine-grained perception. While FineCLIP better captures local semantics than CLIP, it fails to maintain coherent global understanding. In contrast, HarmoCLIP not only accurately identifies multiple salient objects at the global level but also exhibits highly focused token-level attention.
These visualizations clearly demonstrate that HarmoCLIP achieves consistent alignment between textual and visual regions, effectively resolving the long-standing trade-off between global semantic awareness and fine-grained localization in CLIP-based models.

\section{Case Study}

\begin{figure*}[t]
    \centering
    \includegraphics[width=\linewidth, clip]{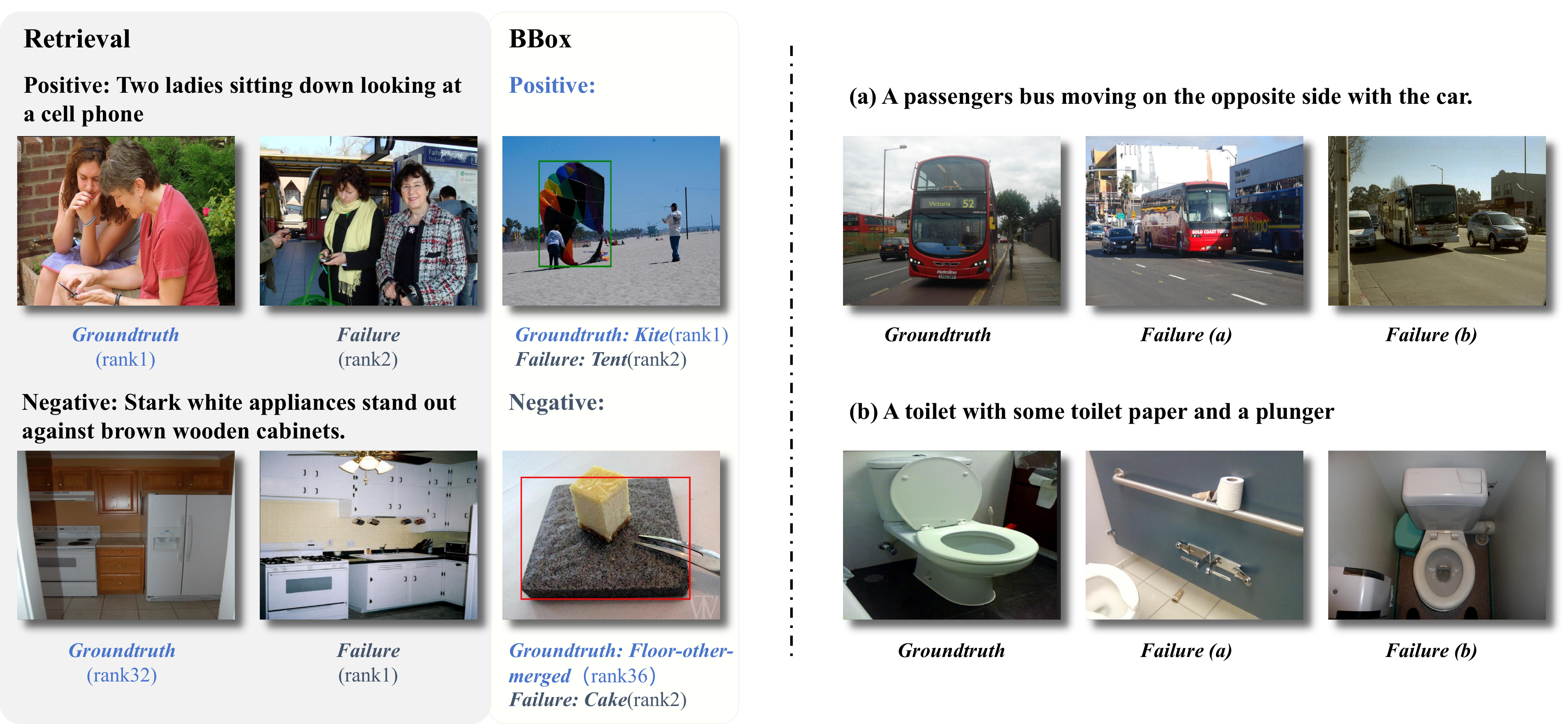}
    \caption{\textbf{Positive and Negative Case in Retrieval or BBox classification}}
    \label{fig:supplementary_4}
\end{figure*}

In Figure \ref{fig:supplementary_4}, we present a series of positive and negative examples from both global-level and region-level tasks to more thoroughly illustrate the effectiveness and limitations of HarmoCLIP.
For the global task retrieval, HarmoCLIP demonstrates strong action-level understanding. In the positive case, where both CLIP and FineCLIP fail to rank the ground truth and mistakenly assign the highest rank to noise samples, HarmoCLIP successfully identifies the correct match. This highlights its superior ability to reason over partially occluded cues and subtle action semantics.
However, in the negative case, HarmoCLIP fails to distinguish the noise sample from the ground truth due to an incorrect inference of object color. This limitation arises because, during training, the masked tokens are restricted to object nouns rather than adjectives; consequently, fine-grained attributes such as color, though implicitly encoded in the hidden states, are not directly supervised.

For the BBox classification task, our method exhibits accurate local object discrimination. Through explicit supervision on the hidden states of object-related tokens, HarmoCLIP leverages both the region itself and surrounding contextual cues to make more reliable predictions. As shown in the positive case, HarmoCLIP correctly identifies the object as kite by incorporating contextual information, whereas competing models incorrectly predict tent.
Nonetheless, HarmoCLIP shows weaker robustness when dealing with background regions, especially those containing large distractor objects. In the negative case, HarmoCLIP predicts cake for a background region because a large portion of cake is visible within the bounding box. The model implicitly attempts to “complete’’ or infer the missing visual information based on nearby evidence—an artifact directly related to our training strategy.


\section{Discussion and Future Work}
In the course of our study, we also uncovered several interesting phenomena that may provide additional insight for the community. As shown in Figure~\ref{fig:supplementary_4} (a), current models find it difficult to correctly interpret relational semantics such as the preposition “opposite.” When objects belong to the same category but differ in attributes such as adjectives or adverbs, the models frequently fail to distinguish the groundtruth image from distractors. We hypothesize that this phenomenon arises because the supervision in our framework focuses primarily on noun-level alignment, without explicitly modeling other parts of speech. Introducing additional supervision for adjectives, adverbs, and even clause-level structures may help the models learn more refined and discriminative semantic cues.

Figure~\ref{fig:supplementary_4}(b) further presents a challenging scenario in which none of the evaluated models successfully retrieve the correct image. The key objects described in the caption, such as \textit{plunger} and \textit{toilet paper}, are extremely small and partially occluded. As a result, all models incorrectly retrieve images that do not contain these objects. This observation indicates that existing CLIP-based approaches lack the ability to perform retrieval reasoning, which involves inferring or localizing objects that are only partially visible in complex scenes. The absence of such reasoning limits the capacity to interpret difficult cases that require contextual imagination or robust scene completion.

For future research, we believe these observations suggest a promising direction involving more structured and fine-grained supervision aimed at enhancing both perceptual discrimination and reasoning ability. We hope these findings provide additional insight for the community and inspire the development of more scalable and fine-grained alignment strategies. 

\end{document}